\documentclass[journal,twoside]{IEEEtran}

\ifCLASSINFOpdf
\else
\fi

\usepackage{algpseudocode}
\usepackage{algorithm}

\usepackage{amsmath,amssymb,amsthm,amsfonts}
\usepackage{breqn}
\usepackage{mathtools}
\usepackage{cases}

\usepackage{accents}
\usepackage{stfloats}
\usepackage{graphicx}
\usepackage{subcaption}
\usepackage{mwe}
\usepackage[export]{adjustbox}
\usepackage{array}

\usepackage{booktabs}
\usepackage{siunitx}
\usepackage[table]{xcolor}

\usepackage{caption}
\usepackage{float}

\usepackage{makecell}
\usepackage{caption}

\usepackage{times}
\newcommand{\subparagraph}{}
\usepackage{titlesec}
\titlespacing\section{0pt}{4pt plus 4pt minus 2pt}{4pt plus 2pt minus 2pt}
\titlespacing\subsection{0pt}{3pt plus 2pt minus 2pt}{3pt plus 2pt minus 2pt}
\usepackage{array}

\usepackage[normalem]{ulem}
\useunder{\uline}{\ul}{}
\usepackage{balance}
\usepackage{epstopdf}
\usepackage{scrextend}
\usepackage{cite}
\usepackage{amsmath}

\usepackage{amsmath}
\usepackage{amssymb}
\usepackage{multirow}
\let\OLDthebibliography\thebibliography
\renewcommand\thebibliography[1]{
  \OLDthebibliography{#1}
  \setlength{\parskip}{0.3pt}
  \setlength{\itemsep}{0.5pt plus 0.6ex}
}
\newlength{\dhatheight}

\newcolumntype{P}[1]{>{\centering\arraybackslash}p{#1}}
\newcolumntype{M}[1]{>{\centering\arraybackslash}m{#1}}
\makeatletter
\newcommand*\bigcdot{\mathpalette\bigcdot@{1}}
\newcommand*\bigcdot@[2]{\mathbin{\vcenter{\hbox{\scalebox{#2}{$\m@th#1\bullet$}}}}}
\makeatother

\begin{document}

\title{NVE: A Separability and Coverage-Aware Internal Validation Metric for Biclustering}
%
%
%

\author{
Paritosh~Tiwari,
I~Navin~Kumar,
James~C.~Bezdek,
and Punit~Rathore
\thanks{P. Tiwari is with the Robert Bosch Center for Cyber-Physical Systems (RBCCPS), Indian Institute of Science, Bengaluru, India (e-mail: paritosht@iisc.ac.in). I. Navin Kumar was with RBCCPS, Indian Institute of Science, Bengaluru, India, at the time of this work. He is now with Cisco India}
\thanks{J. C. Bezdek is a Visiting Senior Research Fellow, University of Melbourne, Australia.}
\thanks{P. Rathore is an Assistant Professor with the Robert Bosch Centre for Cyber-Physical Systems (RBCCPS) and the Centre for Infrastructure, Sustainable Transportation and Urban Planning (CiSTUP), Indian Institute of Science, Bengaluru, India.}
}




\maketitle

\begin{abstract}
Biclustering, or co-clustering, aims to discover coherent submatrices by grouping rows and columns of a data matrix simultaneously. This local two-dimensional structure makes validation more difficult than in ordinary clustering, where internal indices usually rely on compactness and separation in a single shared feature space. Existing popular internal biclustering measures such as Mean Squared Residue (MSR), and Virtual Error (VE) mainly evaluate within-bicluster coherence. Although useful, these measures do not directly assess whether the extracted biclusters are mutually distinct or whether they explain a meaningful portion of the data matrix. This paper investigates Normalised Virtual Error (NVE), an internal validation metric that extends VE using a super-bicluster normalization strategy. By comparing the VE of each bicluster with the VE obtained after merging it with other biclusters, NVE introduces a relative notion of separability and redundancy. We also study a coverage-adjusted variant, NVE\textsubscript{cov}, which penalizes solutions that obtain low error by selecting only very small submatrices. Through controlled synthetic benchmarks and yeast gene-expression datasets, we examine whether NVE and NVE\textsubscript{cov} provide information beyond standard coherence-based metrics. The results show that NVE is sensitive to redundant and poorly separated biclusters, while NVE\textsubscript{cov} changes solution rankings when low-error biclusters cover only a negligible part of the matrix. These findings suggest that NVE-based measures are useful complementary criteria for internal co-clustering validation, especially when coherence, separability, and coverage must be considered jointly.
\end{abstract}
\begin{IEEEkeywords}
Biclustering, co-clustering, internal validation, cluster validity, virtual error, normalized virtual error, mean squared residue, bicluster quality, gene expression analysis.
\end{IEEEkeywords}

\section{Introduction}

\IEEEPARstart{B}{iclustering}, also referred to as co-clustering or two-mode clustering, is used when meaningful structure is expected to appear in both dimensions of a rectangular data matrix. Instead of assigning objects to clusters over the full feature space, biclustering searches for submatrices whose rows exhibit coherent behaviour over selected subsets of columns. This formulation is natural in gene-expression analysis, document-word analysis, recommender systems, and other relational data settings, where a group of objects may be similar only under a limited set of conditions or attributes. Foundational work such as Cheng and Church's residue-based biclustering model \cite{cheng2000biclustering} and Dhillon's bipartite spectral co-clustering formulation \cite{dhillon2001co} established biclustering/co-clustering as a distinct problem rather than a direct extension of ordinary clustering. Subsequent surveys have emphasized the diversity of bicluster types, search strategies, and validation criteria that arise from this local two-dimensional structure \cite{madeira2004biclustering,noronha2022impact,castanho2024survey}.

The evaluation problem is correspondingly different from standard cluster validation. In conventional clustering, many internal validity indices are built around compactness and separation in a single shared feature space. A cluster is good if its points are close to one another and well separated from points in other clusters. For a bicluster, this geometry is no longer sufficient. A bicluster $B_k=(I_k,J_k)$ occupies only the submatrix induced by row set $I_k$ and column set $J_k$; two biclusters may overlap, use different column subsets, or express different coherence models. As a result, a global distance between row clusters or a global variance decomposition does not directly measure whether the extracted submatrices are meaningful. A useful internal validation measure for biclustering must therefore ask a more specific question: \textit{are the reported submatrices internally coherent, mutually non-redundant, and large enough to explain a non-trivial portion of the data matrix}?

Most widely used internal biclustering measures address only the first part of this question. The mean squared residue (MSR) introduced by Cheng and Church evaluates the fit of an additive row-column model within a bicluster \cite{cheng2000biclustering}. This measure is useful, but its interpretation depends on the assumed coherence model. Particularly, residue-based scores may penalize scaling patterns that are structurally meaningful in applications such as gene-expression analysis. Virtual Error (VE) was introduced to address this limitation by comparing standardized row profiles with a virtual pattern representing the average behaviour of the bicluster \cite{divina2012effective}. Comparative studies of bicluster quality measures show that such measures can respond quite differently to constant, shifting, scaling, and combined patterns \cite{pontes2015biclustering}. Thus, internal coherence is not a single universal property; it is tied to the pattern model that the measure is designed to capture.

Even when a coherence measure is appropriate, evaluating each bicluster in isolation leaves two important gaps. First, a collection of biclusters may contain redundant structures. If two biclusters have nearly the same behavioural pattern, merging them may not substantially degrade coherence; a solution composed of such repeated patterns should not be judged as strongly as a solution containing distinct local structures. Second, coherence-only criteria can favour very small submatrices. Small biclusters can appear extremely clean, sometimes for purely numerical or statistical reasons, while covering only a negligible fraction of the matrix. External validation and benchmark-based comparisons can address some of these issues when a reference solution is available \cite{prelic2006comparison,padilha2017systematic}, and statistical significance methods provide another complementary view under a chosen null model \cite{tanay2002discovering,henriques2017bsig}. However, many practical settings require an internal criterion that can be computed directly from the data matrix and the extracted biclusters, without ground truth labels or application-specific annotations.

This paper studies whether our proposed internal validation metric, \textit{Normalised Virtual Error} (NVE), and its coverage-aware variant provide such additional internal validation information. The starting point is the observation that VE captures a behaviourally meaningful form of within-bicluster consistency, but does not by itself compare a bicluster against the rest of the solution. We therefore combine the VE coherence model with a super-bicluster normalization idea related to the normalized squared residue framework of Lee et al. \cite{koreanPaper}. For a bicluster $B_k$, NVE compares its VE with the VE obtained after merging it with alternative biclusters. If $B_k$ is genuinely distinct, these merged super-biclusters should be less coherent; if it is redundant, the normalization exposes that lack of separation. We further consider a coverage-adjusted form, denoted NVE\textsubscript{cov}, which penalizes solutions that achieve low error by explaining only a very small portion of the matrix.

The central question of the paper is therefore not whether NVE should replace all existing biclustering validation measures. Rather, we ask whether NVE contributes information that is not already captured by standard coherence-based measures such as MSR and VE. This distinction is important. A useful validation measure need not be universally best; it should make visible a property of the solution that would otherwise be hidden. In the present case, the properties of interest are behavioural consistency, relative separability, and coverage. These properties are related but not equivalent, and treating them as interchangeable can lead to different conclusions about which biclustering solution is preferable.

The main contributions of this paper are as follows:
\begin{itemize}
    \item We formulate NVE as an internal, algorithm-agnostic validation criterion that extends VE from an isolated bicluster coherence score to a solution-level measure incorporating relative separability through super-biclusters.
    \item We introduce a coverage-adjusted variant, NVE\textsubscript{cov}, to reduce the tendency of coherence-only measures to favour small, highly homogeneous biclusters with limited representativeness.
    \item We design controlled synthetic scenarios that isolate specific evaluation failure modes, including scaling behaviour, redundant or overlapping biclusters, small-clean versus large-noisy structures, thin-column degeneracy, and coverage cherry-picking.
    \item We compare MSR, VE, NVE, and NVE\textsubscript{cov} on both synthetic benchmarks and yeast gene-expression datasets, examining whether the proposed measures alter solution ranking and model selection in practice.
\end{itemize}

Overall, the paper positions NVE as a complementary internal validation measure for co-clustering. Its purpose is to make separability and representativeness visible alongside behavioural coherence, while retaining the pattern sensitivity that motivated VE. The remainder of the paper introduces the necessary background, reviews related biclustering validation work, defines NVE and NVE\textsubscript{cov}, and evaluates their behaviour in controlled and real-data experiments.
\section{Background}
\label{sec:background}

\subsection{Biclustering Overview}
Biclustering, also referred to as co-clustering in parts of the literature, aims to identify localized structure in a data matrix by selecting subsets of rows and columns simultaneously. Let $\widetilde{D}_{m\times n}$ denote the data matrix, where each element $\widetilde{d}_{ij}$ represents the interaction between object $x_i \in X = \{x_1, x_2, \ldots, x_m\}$ and feature $y_j \in Y = \{y_1, y_2, \ldots, y_n\}$. The goal is to obtain a set of $K$ biclusters $\mathcal{O} = \{(O_k, F_k)\}_{k=1}^K$, where $O_k \subseteq X$ and $F_k \subseteq Y$ define subsets of rows and columns such that the corresponding submatrix $B_k = \widetilde{D}(O_k, F_k)$ exhibits high internal coherence and interpretable structure.

Each bicluster $B_k$ thus captures a subset of objects that show a coherent behavioural pattern across a subset of features. This local two-dimensional representation distinguishes biclustering from conventional clustering, which groups objects over the full feature space or partitions along a single axis. Formally, the process can be viewed as finding a mapping
\[
\Phi: \widetilde{D}_{m\times n} \mapsto \{B_1, B_2, \ldots, B_K\},
\]
such that intra-bicluster similarity is maximized while while different biclusters remain meaningfully distinguishable. Owing to this bidirectional structure, validation metrics for biclustering must account for dependencies between the row and column spaces, rather than evaluating each independently.


Several methodological families of biclustering and co-clustering algorithms have been developed, each reflecting different assumptions about the structure of the data matrix. Cheng and Church~\cite{cheng2000biclustering} introduced a residue-based formulation that searches for submatrices with low mean squared residue, while Dhillon~\cite{dhillon2001co} cast co-clustering as bipartite spectral graph partitioning. Large Average Submatrices (LAS)~\cite{shabalin2009finding} instead adopts a statistical perspective, seeking submatrices whose average signal is unusually large relative to the background. These approaches represent complementary residue-based, graph-theoretic, and statistical views of the same broad goal: recovering coherent local submatrices from high-dimensional data. The validation measures studied in this paper are algorithm-agnostic and require only the data matrix and the final collection of biclusters.

\subsection{Bicluster Properties}

Validation in biclustering is more involved than in ordinary clustering because a bicluster is defined jointly by a row subset and a column subset. Thus, the quality of a bicluster cannot be judged only by compactness in a single feature space. Let $A \in \mathbb{R}^{n \times m}$ be the data matrix, and let $B_k=(I_k,J_k)$ denote a bicluster with row set $I_k \subseteq \{1,\dots,n\}$ and column set $J_k \subseteq \{1,\dots,m\}$. A useful validation measure should assess whether the induced submatrix is internally coherent, whether its pattern agrees with the intended bicluster model, and whether the collection of recovered biclusters is informative at the solution level.

Several properties are therefore relevant when evaluating a biclustering solution. First, a bicluster should exhibit strong internal coherence, commonly measured by residue- or error-based criteria such as MSR and VE~\cite{cheng2000biclustering,divina2012effective}. Second, this coherence should correspond to an appropriate pattern model. Depending on the application, meaningful biclusters may be constant, additive, multiplicative, order-preserving, or combinations of these patterns~\cite{madeira2004biclustering, pontes2015biclustering}. Third, biclusters should have non-trivial size and coverage: very small submatrices can appear highly coherent while explaining little of the data matrix. Fourth, at the solution level, the recovered biclusters should not be excessively redundant. Although overlap is natural in biclustering, repeated discovery of nearly identical row--column structures reduces interpretability and adds little new information. Finally, statistical significance and robustness to noise or missing values are also desirable, but they typically require additional null-model, perturbation, or application-specific assumptions~\cite{henriques2017bsig,castanho2024survey}.

This paper does not attempt to optimize all of these properties simultaneously. Instead, they define the evaluation landscape in which Normalised Virtual Error is positioned. The proposed metric family focuses on three internal dimensions that can be computed directly from the data matrix and the extracted biclusters: behavioural coherence through VE, relative distinctness through super-bicluster comparison, and representativeness through union coverage. In this sense, NVE and NVE$_{\mathrm{cov}}$ are not intended to replace external validation, statistical significance analysis, or robustness studies. Their role is narrower: to test whether separability and coverage expose information that standard coherence-only summaries do not capture.

\subsection{Existing Internal Validation Measures}

\subsubsection{Mean Squared Residue (MSR)}

MSR was introduced by Cheng and Church~\cite{cheng2000biclustering} and remains one of the most widely used internal measures of bicluster coherence. For a bicluster $B_k=(I_k,J_k)$ extracted from a data matrix $A$, MSR is defined as
\begin{equation}
\label{eq:msr}
\mathrm{MSR}(B_k)
=
\frac{1}{|I_k||J_k|}
\sum_{i \in I_k}\sum_{j \in J_k}
\left(
a_{ij}-a_{iJ_k}-a_{I_kj}+a_{I_kJ_k}
\right)^2 ,
\end{equation}
where $a_{ij}$ denotes the entry of the data matrix $A$ in row $i$ and column $j$, and
\begin{equation}
\label{eq:msr2}
\begin{aligned}
a_{iJ_k} &= \frac{1}{|J_k|}\sum_{j \in J_k} a_{ij}, \qquad
a_{I_k j} = \frac{1}{|I_k|}\sum_{i \in I_k} a_{ij}, \\
a_{I_k J_k} &= \frac{1}{|I_k||J_k|}
\sum_{i \in I_k}\sum_{j \in J_k} a_{ij}.
\end{aligned}
\end{equation}

Smaller MSR values indicate stronger fit to an additive row--column model. For a biclustering solution $\mathcal{B}=\{B_1,\dots,B_K\}$, the average MSR, also called ASR, is
\begin{equation}
\label{eq:asr}
\mathrm{ASR}(\mathcal{B})
=
\frac{1}{K}
\sum_{k=1}^{K}\mathrm{MSR}(B_k).
\end{equation}

Although MSR is useful and widely adopted, it evaluates only within-bicluster additive coherence. It can therefore penalize meaningful shifting or scaling patterns and does not account for bicluster size, redundancy, or coverage~\cite{divina2012effective,castanho2024survey}. These limitations motivate the use of complementary measures such as VE, NVE, and NVE$_{\mathrm{cov}}$.

\subsubsection{Virtual Error (VE)}

VE was introduced by Divina \textit{et al.} and was proposed to address some of the known limitations of MSR as a bicluster quality measure~\cite{divina2012effective}. In particular, VE is designed to evaluate the consistency of the overall behaviour of the rows in a bicluster with respect to a representative \textit{virtual pattern}, making it more suitable for pattern-based biclusters.

For a bicluster $B_k=(I_k,J_k)$, the virtual pattern is defined over the selected columns and is given by
\begin{equation}
r_j = \frac{1}{|I_k|}\sum_{i\in I_k} b_{ij}, \qquad j\in J_k,
\end{equation}
where $b_{ij}$ denotes the entry of the bicluster submatrix at row $i$ and column $j$. Thus, the virtual pattern summarizes the common trend of the bicluster across its columns.

To focus on behaviour rather than absolute magnitude, both the row profiles of the bicluster and the virtual pattern are standardized before comparison~\cite{colantuoni2002snomad}. Let $\hat{b}_{ij}$ denote the standardized value of entry $b_{ij}$ and let $\hat{r}_j$ denote the standardized value of the corresponding component of the virtual pattern. The VE of bicluster $B_k$ is then defined as
\begin{equation}
\label{eq:ve}
\mathrm{VE}(B_k)=\frac{1}{|I_k||J_k|}\sum_{i\in I_k}\sum_{j\in J_k}\left|\hat{b}_{ij}-\hat{r}_j\right|.
\end{equation}

Similar to Equation~\ref{eq:asr}, we can report the average VE across all discovered biclusters for a biclustering solution:
\begin{equation}
\mathrm{AvgVE}(\mathcal{B})=\frac{1}{K}\sum_{k=1}^{K}\mathrm{VE}(B_k).
\end{equation}

Lower VE values indicate that the rows in the bicluster follow a common standardized pattern more closely. For this reason, VE is better suited than MSR for assessing biclusters that exhibit coherent behavioural trends, including shifting and scaling patterns~\cite{divina2012effective}. However, VE remains an internal within-bicluster coherence measure. It evaluates how well the entries of a bicluster agree with their virtual pattern, but it does not directly account for between-bicluster distinctness, overlap, or redundancy within the full biclustering solution.

\subsubsection{Normalised Squared Residue (NSR)}

NSR, proposed by Lee \textit{et al.}~\cite{koreanPaper}, extends MSR by incorporating a relative notion of distinctness between biclusters. Instead of evaluating each bicluster only in isolation, NSR compares its internal coherence with the coherence of a larger \textit{super bicluster} formed by merging it with another bicluster.

Let a biclustering solution be $\mathcal{B}=\{B_1,\dots,B_K\}$, where each bicluster is written as $B_k=(I_k,J_k)$. For two biclusters $B_k$ and $B_\ell$, their super bicluster is defined as
\begin{equation}
B_{k\ell}=(I_k \cup I_\ell,\; J_k \cup J_\ell).
\end{equation}
The quantity $\mathrm{MSR}(B_{k\ell})$, as computed in Eq.~\eqref{eq:msr}, then measures the residue of the merged structure. If two biclusters are genuinely distinct, their merger is expected to be less coherent, leading to a larger MSR.

A solution-level form of NSR can therefore be written as
\begin{equation}
\label{eq:nsr}
\mathrm{NSR}(\mathcal{B})
=
\frac{\sum_{k=1}^{K}\mathrm{MSR}(B_k)}
{\sum_{k=1}^{K}\min_{\ell \neq k}\mathrm{MSR}(B_{k\ell})}.
\end{equation}
Lower values indicate better quality, since they reflect low within-bicluster residue together with a stronger degradation of coherence when biclusters are merged with their closest counterparts. Thus, unlike MSR, NSR attempts to assess not only internal homogeneity but also the relative distinctness of biclusters. However, because it is still built on MSR, it inherits MSR's limitations in handling broader pattern-based structures, particularly shifting and scaling relationships~\cite{divina2012effective}.

\subsection{Related Work}
\label{subsec:related_work_positioning}

The preceding sections define the biclustering setting and the internal measures used in this study. We therefore use the related work mainly to delimit the evaluation question addressed here. The residue-based model of Cheng and Church and the bipartite spectral formulation of Dhillon remain two canonical starting points for biclustering and co-clustering, while the survey of Madeira and Oliveira organized the field in terms of bicluster types, algorithmic strategies, and evaluation criteria \cite{cheng2000biclustering,dhillon2001co,madeira2004biclustering}. More recent surveys make the same distinction explicit: biclustering evaluation includes internal quality, external recovery, statistical significance, visualization, and application-dependent validation, and these should not be treated as interchangeable objectives \cite{noronha2022impact,castanho2024survey}.

Prior work on internal validity measures shows why the choice of coherence model matters. Aguilar-Ruiz formalized shifting and scaling patterns as meaningful structures in gene-expression data \cite{aguilar2005shifting}, and Divina et al. proposed Virtual Error (VE) to evaluate standardized behavioural consistency rather than only additive residue \cite{divina2012effective}. Pontes et al. compared a broad family of bicluster quality measures and showed that different indices respond differently to constant, shifting, scaling, and combined patterns \cite{pontes2015biclustering}. These results are central to the motivation of the present paper: MSR/ASR and VE/AvgVE are informative reference axes, but they do not by themselves measure whether the biclusters in a full solution are mutually distinct or whether the solution covers a meaningful part of the matrix.

External validation and benchmark-based comparisons address a different problem. Prelic et al. proposed an influential experimental framework combining synthetic recovery and biological validation \cite{prelic2006comparison}, while Eren et al. emphasized that comparisons based on a single planted model can be misleading because algorithms optimize different bicluster models \cite{eren2013comparative}. Horta and Campello studied desirable properties of external similarity measures for comparing biclustering solutions \cite{horta2014similarity}, and Padilha and Campello extended this direction in a large comparative evaluation using synthetic and biological evidence \cite{padilha2017systematic}. Benchmark generators such as G-Bic further support controlled assessment under varied coherence, overlap, and noise conditions \cite{castanho2023gbic}. These studies are relevant to experimental design, but their validation setting is not the same as ours: external measures require a planted or reference solution, whereas NVE is intended for internal comparison when only the data matrix and the extracted biclusters are available.

Statistical significance provides another complementary view. Early graph-theoretic approaches searched for biclusters unlikely to arise under a random model, and BSig later evaluated the statistical significance of biclustering solutions under different coherence assumptions \cite{tanay2002discovering,henriques2017bsig}. Pattern mining-based biclustering also makes the relation between coherence assumptions, search constraints, and quality criteria explicit \cite{henriques2015structured}. Such methods are important because small coherent biclusters can occur by chance in high-dimensional data. However, significance testing depends on a specified null model and does not directly provide the simple solution-level comparison targeted here, namely a deterministic score combining behavioural coherence, relative separability, and coverage.

The closest prior internal criterion is the normalized squared residue (NSR) framework of Lee et al. \cite{koreanPaper}. NSR uses a super-bicluster comparison idea: if two biclusters represent distinct structures, merging them should degrade coherence. NVE keeps this solution-level normalization principle but replaces the additive-residue model with the VE behavioural model. The coverage-adjusted variant then adds an explicit representativeness term so that a solution formed from very small, clean biclusters is not automatically preferred over a broader, moderately noisy solution. Thus, the contribution of NVE is not to replace external validation, significance analysis, or all existing internal indices. Its narrower role is to test whether VE-based coherence, super-bicluster separability, and union coverage expose information that standard coherence summaries do not capture.

\section{Normalised Virtual Error}

The internal validation measures reviewed above capture different aspects of bicluster quality, but none of them provides a complete assessment on its own. In particular, standard coherence measures such as MSR and VE focus on the internal consistency of each bicluster, but they do not directly reward the recovery of larger and more informative submatrices. As a result, relying on coherence alone can bias evaluation toward small biclusters, even when larger biclusters reveal broader and more useful structure in the data. In biclustering, this is an important limitation, since a good solution should ideally balance internal coherence with non-trivial size.

A second limitation is that most internal measures evaluate biclusters in isolation. Among the measures discussed earlier, only NSR incorporates a relative notion of distinctness by comparing a bicluster with merged alternatives. This is useful because a high-quality biclustering solution should contain biclusters that are not only internally coherent, but also meaningfully distinguishable from one another. If merging two biclusters substantially degrades their coherence, this suggests that they represent separate structures rather than redundant variations of the same pattern.

To address these limitations, we propose the \emph{Normalised Virtual Error} (NVE), which extends VE in two ways. First, it incorporates a normalization strategy inspired by NSR, so that the coherence of each bicluster is assessed relative to the coherence of merged alternatives. Second, it can be coupled with an explicit size or coverage factor to account for the area of the matrix captured by the bicluster. In this way, NVE is intended as an algorithm-agnostic internal evaluation measure for biclustering and co-clustering solutions.

\subsection{NVE Definition}

Let $\mathcal{B}=\{B_1,\dots,B_K\}$ be a biclustering solution on a data matrix $A \in \mathbb{R}^{n\times m}$, where each bicluster is written as $B_k=(I_k,J_k)$ with row index set $I_k$ and column index set $J_k$. For two biclusters $B_k$ and $B_j$, define the corresponding super bicluster as $B_{kj} = (I_k \cup I_j,\; J_k \cup J_j)$.

Using the Virtual Error $\mathrm{VE}(B_k)$ of each, as in Eq.~\eqref{eq:ve}, we first define the normalized contribution of $B_k$ as
\begin{equation}
\label{eq:nve-single}
\mathrm{NVE}_k
=
\frac{\mathrm{VE}(B_k)}
{\min_{j \neq k}\mathrm{VE}(B_{kj})}.
\end{equation}
The denominator compares $B_k$ against the most coherent super bicluster obtained by merging it with another bicluster. When $B_k$ is genuinely distinct, merging it with any other bicluster is expected to increase the virtual error, leading to a small normalized ratio.

The overall NVE of the biclustering solution is then defined as
\begin{equation}
\label{eq:nve-sum}
\mathrm{NVE}(\mathcal{B})
=
\frac{\sum_{k=1}^{K}\mathrm{VE}(B_k)}
{\sum_{k=1}^{K}\min_{j \neq k}\mathrm{VE}(B_{kj})}.
\end{equation}
Lower values of $\mathrm{NVE}(\mathcal{B})$ indicate better solutions, as they correspond to biclusters that are internally pattern-consistent while remaining relatively distinct from their merged alternatives.

In this form, NVE remains a general internal validation measure: it does not depend on how the biclusters were generated, and it can therefore be applied to the outputs of different biclustering or co-clustering algorithms in a uniform way.

\subsection{Coverage-adjusted Normalised Virtual Error.}


The vanilla Normalised Virtual Error (NVE) evaluates a biclustering solution by combining within-bicluster behavioural coherence, captured through VE, with a normalization term based on super-biclusters. In this way, NVE reflects both internal pattern consistency and a relative notion of distinctness. However, vanilla NVE does not explicitly reward the size or representativeness of the extracted biclusters. Consequently, a solution made up of a few very small but internally coherent biclusters may still obtain a favorable score even when it explains only a negligible fraction of the data matrix. This is an important limitation in biclustering, where the size, overlap, and overall coverage of the recovered biclusters are widely recognized as relevant structural aspects of a solution~\cite{castanho2024survey,waltman2010multispecies,sutheeworapong2012novel}.

To address this limitation, we introduce a coverage-adjusted
variant of NVE, denoted by \(\mathrm{NVE}_{\mathrm{cov}}\), defined as
\begin{equation}
\mathrm{NVE}_{\mathrm{cov}}(\mathcal{B})
=
\frac{\mathrm{NVE}(\mathcal{B})}
{\left(\mathrm{CR}(\mathcal{B})+\epsilon\right)^{1/2}},
\label{eq:nve_cov}
\end{equation}
where \(\mathcal{B}=\{B_1,\ldots,B_K\}\) is the set of extracted biclusters, \(\mathrm{NVE}(\mathcal{B})\) is the vanilla NVE score,
\(\mathrm{CR}(\mathcal{B})\in[0,1]\) is the coverage ratio of the solution, and \(\epsilon>0\) is a small numerical constant introduced for stability. The coverage ratio is defined as
\begin{equation}
\Omega(B_k)=I_k\times J_k,
\qquad
\mathrm{CR}(\mathcal{B})
=
\frac{\left|\bigcup_{k=1}^{K}\Omega(B_k)\right|}{nm}.
\label{eq:coverage_ratio}
\end{equation}
Here, \(\Omega(B_k)\) denotes the set of matrix positions covered by bicluster \(B_k=(I_k,J_k)\). Thus,
\(\bigcup_{k=1}^{K}\Omega(B_k)\) is the set of distinct entries covered by at least one bicluster in the solution, and \(nm\) is the total number of entries in the data matrix \(A\in\mathbb{R}^{n\times m}\).

The proposed adjustment preserves the original interpretation of NVE as a minimization criterion, while introducing a global representativeness term at the solution level. When the union coverage is small, the denominator in Eq.~\eqref{eq:nve_cov} decreases and the score is penalized accordingly. Conversely, among solutions with comparable NVE values, those covering a larger fraction of the matrix receive a lower NVE\textsubscript{cov} score. The exponent $1/2$ provides a moderate adjustment, so that coverage complements rather than overwhelms the coherence and distinctness information already captured by vanilla NVE.


An important advantage of defining \(\mathrm{CR}(\mathcal{B})\) through the union of covered matrix entries, rather than through the mean bicluster area ratio, is that the coverage ratio measures the effective portion of the matrix explained by the full solution. As a result, heavily overlapping biclusters do not receive artificial credit for repeatedly covering the same entries.
This is especially desirable because overlap and coverage are recognized as distinct structural dimensions of biclustering solutions, and excessive overlap can reduce the interpretability of the recovered patterns~\cite{waltman2010multispecies,sutheeworapong2012novel}. In this sense, NVE\textsubscript{cov} complements vanilla NVE by adding a global representativeness criterion: while NVE evaluates coherence and relative distinctness, NVE\textsubscript{cov} additionally evaluates whether the extracted biclusters account for a meaningful portion of the data matrix.
\section{Experiments and Results}

This section examines whether Normalised Virtual Error (NVE) and its coverage-aware variant, NVE\textsubscript{cov}, provide useful evaluation information that is not already captured by existing biclustering metrics. In particular, we want to see whether NVE adds information about how well biclusters are separated from one another beyond what coherence-based measures such as ASR and VE can show. We also want to test whether NVE\textsubscript{cov} adds a meaningful notion of coverage, so that solutions made up of very small but clean biclusters are not always preferred over solutions that explain a larger part of the data. To study these questions, we first use controlled synthetic datasets where the planted structure is known, and then we use real yeast benchmark datasets to see how metric choice affects the ranking and selection of biclustering solutions in practice.

\subsection{Synthetic Benchmark Datasets}
\label{sec:synthetic}

\begin{figure}
  \centering
  \includegraphics[width=\columnwidth]{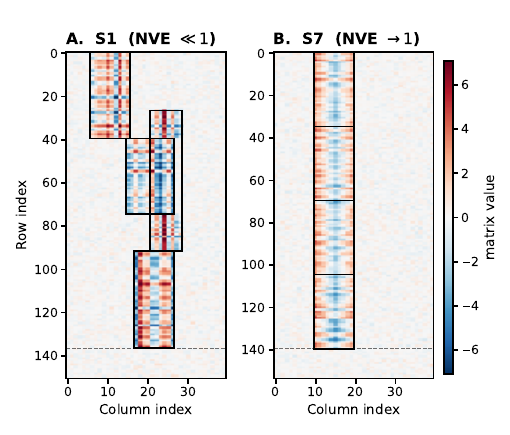}
  \caption{Heatmaps of two synthetic data matrices from Table~\ref{tab:synthetic_scenarios}; rows are permuted so the four planted biclusters $B_1$--$B_4$ stack at the top (matrix values unchanged). Black rectangles mark each bicluster's row-band and column support. In S1 the blocks have distinct column supports and distinct patterns (NVE\,$\ll 1$); in S7 they share the same columns and the same pattern, differing only in per-row shifts, so any pairwise merge reproduces the pattern and NVE\,$\to 1$ even though ASR and VE rate each block individually as good.}
  \label{fig:s1_s7_visual}
\end{figure}

To isolate the information captured by MSR, VE, NVE, and NVE\textsubscript{cov}, we constructed ten controlled synthetic datasets, each represented by a $300 \times 40$ matrix with planted biclusters over a Gaussian background. The dimensions were chosen to be biologically plausible while keeping column overlap non-trivial and computation manageable. Unless overlap is part of the intended stress test, planted biclusters are non-overlapping in the row--column subspace. Table~\ref{tab:synthetic_scenarios} summarises the scenarios and the metric tension each one is designed to expose.

The scenarios are grouped around four diagnostic questions. S1--S3 examine coherence-model dependence: S1 is an additive baseline where MSR, VE, NVE, and NVE\textsubscript{cov} should largely agree; S2 introduces pure scaling, where MSR is expected to penalise a pattern that VE can still regard as coherent; and S3 combines shifting and scaling with mixed-sign patterns, making both residue and virtual-error coherence more difficult while preserving separability. S5 and S7 focus on redundancy and separation. In S5, overlapping biclusters generated from the same underlying pattern should remain favourable under within-bicluster scores but become less attractive under NVE, while S7 is an extreme clone case in which individually coherent biclusters carry little separable information; Fig.~\ref{fig:s1_s7_visual} illustrates this contrast. S4, S9a, and S9b test the coverage--coherence trade-off: small clean biclusters can obtain low error, but NVE\textsubscript{cov} should favour broader solutions when they explain a more meaningful portion of the matrix. Finally, S6 and S8 are diagnostic stress tests. S6 exposes the thin-column degeneracy of VE-based quantities, motivating minimum-size checks, whereas S8 provides a progressive noise ladder for assessing whether metric values respond monotonically to controlled degradation.

This design makes the synthetic benchmark more than a collection of artificial examples. Each scenario creates a known structural condition under which two or more metrics are expected to agree or disagree. The resulting comparisons therefore test whether NVE and NVE\textsubscript{cov} add separability and coverage information beyond standard within-bicluster coherence.

\begin{table*}[t]
\centering
\caption{Summary of synthetic benchmark scenarios. $K$ denotes the number of planted biclusters. The ``Stress test'' column names the metric property or inter-metric tension each scenario is designed to probe.}
\label{tab:synthetic_scenarios}
\resizebox{\textwidth}{!}{%
\begin{tabular}{llcll}
\toprule
\textbf{ID} & \textbf{Name} & $K$ & \textbf{Signal type} & \textbf{Stress test} \\
\midrule
S1  & Additive            & 4 & Pure additive, low noise      & Baseline agreement; NVE $\ll 1$ \\
S2  & Scaling             & 4 & All-positive scaling          & MSR high, VE $\approx 0$; MSR--VE disagreement \\
S3  & Mixed shift-scale   & 4 & Shift + scale, mixed-sign     & Both MSR and VE degrade; NVE separability persists \\
S4  & Size vs.\ quality   & 4 & Additive (2 clean, 2 noisy)   & Coverage--coherence trade-off; NVE\textsubscript{cov} sensitivity \\
S5  & Overlap             & 4 & Additive, equal signal        & Redundancy detection; MSR/VE blind, NVE sensitive \\
S6  & Thin columns        & 4 & Additive + constant stripes   & VE degeneracy; minimum-size filter necessity \\
S7  & Identical clones    & 4 & Same pattern, shared columns  & Zero separability; NVE $\to 1$ despite low MSR/VE \\
S8  & Noise ladder        & 5 & Additive, $\sigma$ increasing & Metric calibration vs.\ controlled quality gradient \\
S9a & Coverage tiling     & 8 & Additive, moderate noise      & NVE\textsubscript{cov} rewards broad coverage \\
S9b & Coverage cherry-pick& 3 & Additive, near-perfect        & NVE\textsubscript{cov} penalises negligible coverage \\
\bottomrule
\end{tabular}%
}
\end{table*}

\subsection{Real Datasets}
We also evaluate the proposed measures on two \textit{Saccharomyces cerevisiae} gene-expression benchmarks that have been used repeatedly in the biclustering literature: the \texttt{alpha\_factor} experiment derived from the \textit{Spellman et al.} cell-cycle dataset, and the \texttt{heat\_shock\_1} experiment derived from the \textit{Gasch et al.} environmental stress-response dataset. The \texttt{alpha\_factor} data capture transcriptional activity across cell-cycle conditions and are known to contain localized groups of genes with coordinated temporal behaviour, whereas \texttt{heat\_shock\_1} reflects the yeast transcriptional response under heat stress and exhibits condition-specific local co-expression patterns. These datasets are complementary: the former emphasizes structured periodic regulation, while the latter reflects stress-induced local responses. Their continued use in prior biclustering studies makes them suitable test beds for examining whether internal validity measures such as VE, NVE, and NVE\textsubscript{cov} distinguish coherent and informative biclustering solutions across different biological regimes.

\subsection{Evaluation of Biclustering Algorithms}
We evaluated three biclustering algorithms in our experiments: Cheng and Church (CCA)~\cite{cheng2000biclustering}, Large Average Submatrices (LAS)~\cite{shabalin2009finding}, and Spectral Biclustering~\cite{kluger2003spectral}. CCA is a residue-based method that searches for submatrices with low mean squared residue, and it serves as a standard coherence-oriented baseline. LAS searches for submatrices with large average values by using repeated randomized searches followed by refinement, and it provides a useful contrast because it often favours strong local patterns. Spectral Biclustering is a matrix decomposition based method that groups rows and columns jointly, giving a third type of biclustering solution that is different from the other two approaches. These three algorithms were chosen because they represent different styles of biclustering and therefore make it easier to see when changes in evaluation are due to the metric rather than to one particular algorithm family. CCA and LAS were implemented as per \cite{padilha2017systematic} and their publicly available library, while Spectral Biclustering used the standard Python implementation corresponding to~\cite{kluger2003spectral}. The outputs of all three algorithms were converted to a common bicluster representation before computing the evaluation measures. In all experiments, the algorithms were run through the same experimental framework, and apart from the requested number of biclusters, their default settings were kept unchanged.

\subsection{Experimental protocol}

The experiments were carried out in two stages. First, we used controlled synthetic datasets in which the planted bicluster structure is known. For these datasets, each algorithm was asked to return the planted number of biclusters for that scenario, so the comparison focuses on how the evaluation metrics behave when the target number of biclusters is fixed. Second, we used the yeast benchmark datasets, where the true bicluster structure is not known in advance. In this setting, each algorithm was run over a small sweep of requested bicluster numbers, with $K \in \{3,5,8,10,15\}$ chosen as a coarse sweep from small to relatively large biclustering solutions. For every resulting solution, we computed ASR, AvgVE, NVE, NVE\textsubscript{cov}, and coverage. The score-based metrics were interpreted in a lower-is-better sense, while coverage was reported separately to show how much of the data matrix was explained by the biclustering solution. Using the same algorithms and the same evaluation metrics in both the synthetic and the real-data experiments allows the results to be compared directly across the two settings.

\subsection{Controlled synthetic validation}

We first used controlled synthetic datasets to check what information the different metrics are actually capturing. These datasets are useful because each one was designed to create a specific evaluation situation, such as clean additive structure, scaling structure, overlap, redundancy, or low coverage. This lets us test the behavior of the metrics in settings where the intended bicluster structure is known in advance. Table~\ref{tab:synthetic_winner_counts} summarizes the algorithm that most often obtains the best score under each metric across the synthetic scenarios. It shows that the preferred algorithm changes with the evaluation criterion. ASR, AvgVE, NVE, and NVE\textsubscript{cov} emphasize different aspects of the output: additive residue, virtual-error coherence, relative separation, and coverage-adjusted separation.

\subsubsection{Coherence-oriented cases.}
The first group of synthetic datasets, namely S1--S3, was used to examine cases where the planted biclusters are well separated but differ in their internal signal model. In S1, the biclusters follow a simple additive pattern, so coherence-based metrics and the NVE family both behave as expected. In S2 and S3, the patterns become more difficult because scaling and mixed shift-scale effects are introduced. These cases are useful because they show that ASR and AvgVE do not always respond in the same way to the same structure. At the same time, the planted biclusters are still distinct from one another, so NVE remains mainly a measure of separability rather than just another measure of internal coherence.

\subsubsection{Separation and redundancy.}
\begin{figure}
    \centering
    \begin{subfigure}{\columnwidth}
        \centering
        \includegraphics[width=\linewidth]{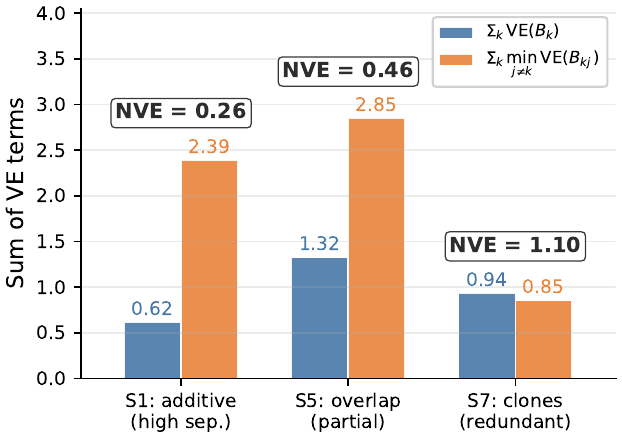}
        \caption{NVE decomposition: numerator vs. denominator}
        \label{fig:nve_decomposition}
    \end{subfigure}
    
    \vspace{0.5em}
    
    \begin{subfigure}{\columnwidth}
        \centering
        \includegraphics[width=\linewidth]{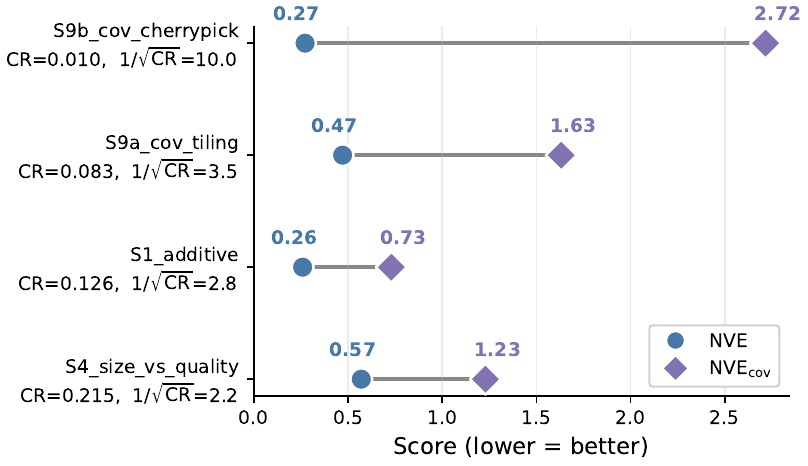}
        \caption{Coverage penalty in NVE\textsubscript{cov} on synthetic data. Here, $\mathrm{CR}$ denotes the coverage ratio defined in Eq.~\eqref{eq:coverage_ratio}.}
        \label{fig:nvecov_penalty}
    \end{subfigure}
    
    \caption{NVE decomposition and coverage penalisation on illustrative synthetic cases.}
    \label{fig:nve_decomp_cov_penalty}
\end{figure}

The separation-specific role of NVE is most transparent in Figure~\ref{fig:nve_decomp_cov_penalty}(\subref{fig:nve_decomposition}). Across S1, S5, and S7, the numerator in Eq.~\eqref{eq:nve-sum}, $\sum_k \mathrm{VE}(B_k)$, remains of the same order but the denominator $\sum_k \min_{j \neq k}\mathrm{VE}(B_{kj})$ changes sharply. In S1, merging two planted biclusters produces a much less coherent super-bicluster, so the denominator is large and NVE is low. In S5, this contrast weakens because overlapping biclusters remain partially compatible after merging. In S7, it nearly disappears: clone biclusters can be merged with little additional incoherence, so NVE moves toward one even though the individual biclusters still appear clean under coherence-only scores. This is precisely the kind of redundancy penalty that ASR and AvgVE do not provide.


\subsubsection{Coverage-related cases.}
The coverage effect is equally explicit in Figure~\ref{fig:nvecov_penalty}. S9b attains the best NVE because the cherry-picked biclusters are nearly perfect internally, but its union coverage is only about $1\%$, so NVE\textsubscript{cov} increases sharply after the coverage penalty is applied. By contrast, S9a and S4 retain slightly worse NVE values yet improve relative to S9b once the solution covers a non-trivial fraction of the matrix. Referencing this plot directly makes the intended distinction clearer: NVE\textsubscript{cov} is not simply a noisier version of NVE, but a criterion that changes the ordering when apparent quality is obtained by explaining too little of the matrix.


Overall, the controlled synthetic experiments serve as a sanity check before moving to algorithm-level comparisons. They show that the proposed metrics were designed to respond to specific structural properties of biclustering solutions: NVE responds to separation and redundancy, while NVE\textsubscript{cov} adds a direct preference against extremely small, low-coverage solutions. This gives a clear basis for the larger synthetic and real-data experiments that follow.

\subsection{Synthetic benchmark results at matched $K$}

After checking the basic behavior of the metrics on planted synthetic structures, we next compared the three algorithms on the full set of synthetic datasets using the planted number of biclusters for each case. This setting is useful because it removes model selection from the comparison. The main question here is not which algorithm can guess the correct number of biclusters, but rather how the different evaluation metrics judge the solutions when the target $K$ is fixed.

\begin{figure}
    \centering
    
    \begin{subfigure}{0.84\columnwidth}
        \includegraphics[width=\linewidth]{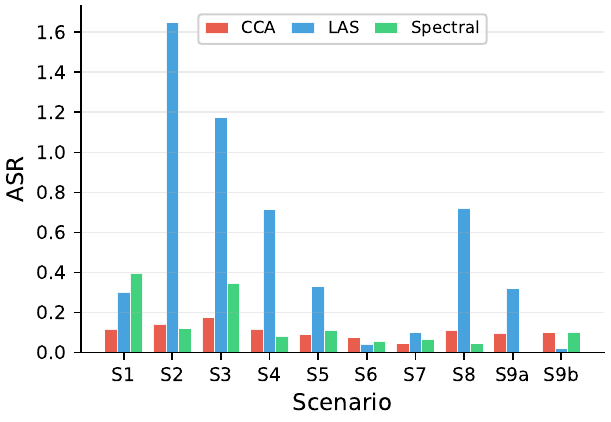}
        \caption{ASR}
        \label{fig:synth_asr}
    \end{subfigure}
    \vspace{0.3em}
    
    \begin{subfigure}{0.84\columnwidth}
        \includegraphics[width=\linewidth]{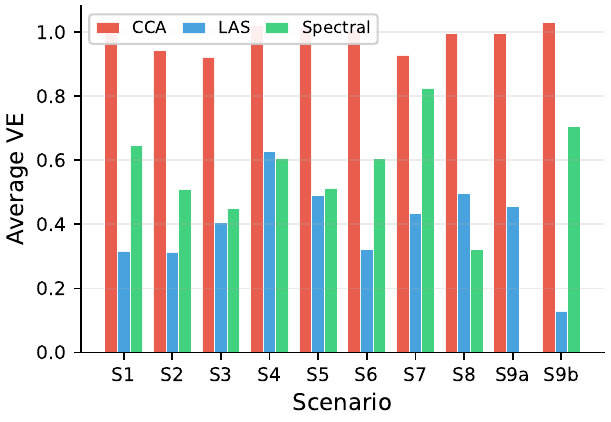}
        \caption{AvgVE}
        \label{fig:synth_avgve}
    \end{subfigure}
    \vspace{0.3em}
    
    \begin{subfigure}{0.84\columnwidth}
        \includegraphics[width=\linewidth]{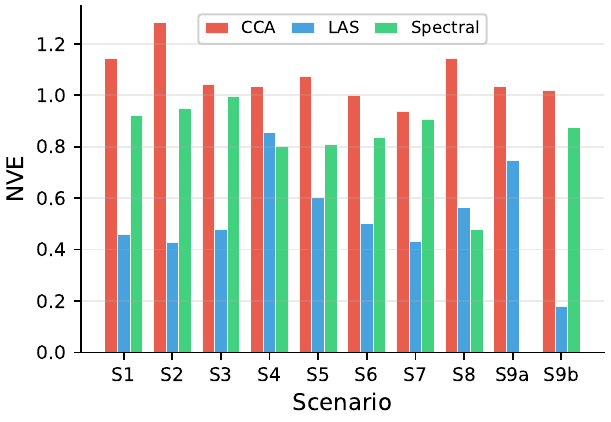}
        \caption{NVE}
        \label{fig:synth_nve}
    \end{subfigure}
    \vspace{0.3em}
    
    \begin{subfigure}{0.84\columnwidth}
        \includegraphics[width=\linewidth]{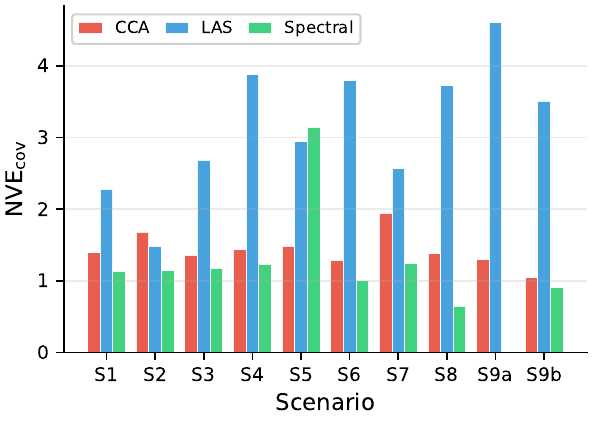}
        \caption{NVE\textsubscript{cov}}
        \label{fig:synth_nvecov}
    \end{subfigure}
    
    \caption{Matched-$K$ synthetic benchmark scores under ASR, AvgVE, NVE, and NVE\textsubscript{cov}. All four metrics are min-optimal.}
    \label{fig:match-k-synth}
\end{figure}

The matched-$K$ synthetic summaries are more informative when the metric-specific plots are cited separately; see Figs.~\ref{fig:match-k-synth}(\subref{fig:synth_asr})--\ref{fig:match-k-synth}(\subref{fig:synth_nvecov}). Figure~\ref{fig:match-k-synth}(\subref{fig:synth_asr}) gives a mixed picture, with CCA or Spectral often preferred depending on the scenario. In contrast, Figs.~\ref{fig:match-k-synth}(\subref{fig:synth_avgve}) and \ref{fig:match-k-synth}(\subref{fig:synth_nve}) favour LAS on most datasets, which is consistent with LAS returning small, locally clean biclusters. The reversal in Fig.~\ref{fig:match-k-synth}(\subref{fig:synth_nvecov}) is therefore substantive rather than cosmetic: once coverage is penalised, Spectral becomes the most frequent winner.


\begin{table}
\centering
\caption{Most frequent winning algorithm under each metric on the synthetic datasets.}
\label{tab:synthetic_winner_counts}
\begin{tabular}{lc}
\hline
Metric & Most frequent winner \\
\hline
ASR & CCA (5/10 datasets) \\
AvgVE & LAS (8/10 datasets) \\
NVE & LAS (8/10 datasets) \\
NVE\textsubscript{cov} & Spectral (8/10 datasets) \\
\hline
\end{tabular}
\end{table}

This shift is especially clear on S2 scaling and S9b coverage cherry-pick. On S2, LAS gives the best AvgVE ($0.313$) and NVE ($0.428$), but it covers only $8.3\%$ of the matrix, whereas Spectral covers $69.3\%$ and becomes best under NVE\textsubscript{cov} ($1.141$). On S9b, LAS again dominates ASR, AvgVE, and NVE, yet its solution covers only $0.27\%$ of the matrix; once coverage is taken into account, Fig.~\ref{fig:match-k-synth}(\subref{fig:synth_nvecov}) shows that the preferred solution switches to Spectral. These examples strengthen the main conclusion of the synthetic benchmark: NVE and especially NVE\textsubscript{cov} do not merely rescale coherence, but can reverse the ranking by penalising narrow, cherry-picked structure.


Taken together, these matched-$K$ synthetic experiments show that metric choice is not a minor detail. If the evaluation is based only on coherence-oriented measures, one family of solutions may look clearly preferable. If separability and coverage are also taken into account, the ranking can change. This is the main result of the synthetic benchmark comparison: NVE and especially NVE\textsubscript{cov} do not simply repeat the information already given by ASR and AvgVE, but can lead to meaningfully different judgments about biclustering quality.

\subsection{Yeast benchmark results under $K$-sweep}

The real-data yeast experiments were used to study a more practical setting in which the true number of biclusters is not known in advance. Instead of fixing $K$ to a planted value, each algorithm was run over a small range of requested bicluster numbers, and the resulting solutions were evaluated by ASR, AvgVE, NVE, and NVE\textsubscript{cov}. This setup is useful because it shows whether the choice of metric affects not only the ranking of algorithms, but also the choice of model complexity. Table~\ref{tab:yeast_k_examples} lists representative yeast cases in which different metrics select different values of $K$ for the same dataset and algorithm. Each entry is the value of $K$ that minimizes the corresponding metric over the tested sweep.

The main result is that the selected solution often changes when the metric changes. Across the yeast experiments, the best $K$ chosen by NVE\textsubscript{cov} differed from the best $K$ chosen by ASR in many dataset-algorithm combinations. Similar differences were also seen when NVE\textsubscript{cov} was compared with AvgVE and NVE. This means that the metric is not only scoring the same solutions in a slightly different way; it is often pointing to a different part of the $K$-sweep altogether. In that sense, the yeast experiments provide stronger evidence than the synthetic matched-$K$ experiments, because they show that metric choice can affect the final model that would actually be selected in practice.

\begin{table}[t]
\centering
\caption{Chosen values of $K$ in the yeast $K$-sweep experiments.}
\label{tab:yeast_k_examples}
\begin{tabular}{lllll}
\hline
Dataset / Algorithm & ASR & AvgVE & NVE & NVE\textsubscript{cov} \\
\hline
01\_alpha\_factor / CCA & 15 & 5 & 5 & 8 \\
02\_cdc\_15 / CCA & 15 & 3 & 3 & 15 \\
02\_cdc\_15 / Spectral & 5 & 5 & 5 & 3 \\
03\_cdc\_28 / CCA & 10 & 8 & 5 & 15 \\
04\_elutriation / CCA & 10 & 5 & 10 & 8 \\
14\_nitrogen\_depletion / CCA & 3 & 3 & 3 & 8 \\
\hline
\end{tabular}
\end{table}

In terms of algorithm ranking, NVE\textsubscript{cov} most often favoured Spectral on the yeast data, while LAS was selected much less often. This is consistent with what was already seen in the synthetic experiments. LAS can obtain very strong values under coherence-oriented criteria, but these solutions often cover only a very small part of the matrix. By contrast, Spectral more often produces solutions with broader coverage, so once the coverage penalty is taken into account, its ranking improves. CCA remains competitive on some datasets, but in general the yeast results suggest that NVE\textsubscript{cov} prefers solutions that strike a better balance between structure quality and extent of coverage.

\begin{figure*}[t]
    \centering
    
    \begin{subfigure}{0.24\textwidth}
        \includegraphics[width=\linewidth]{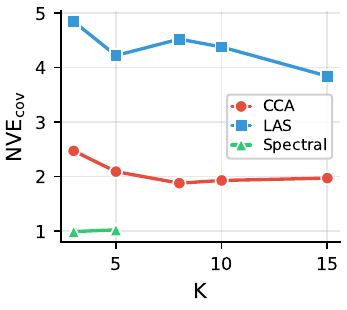}
        \caption{\texttt{\textbf{01_alpha_factor}}}
    \end{subfigure}\hfill
    \begin{subfigure}{0.24\textwidth}
        \includegraphics[width=\linewidth]{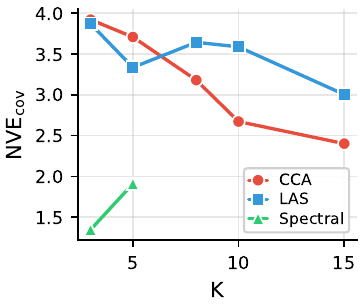}
        \caption{\texttt{\textbf{02_cdc_15}}}
        \label{fig:ksweep_cdc15}
    \end{subfigure}\hfill
    \begin{subfigure}{0.24\textwidth}
        \includegraphics[width=\linewidth]{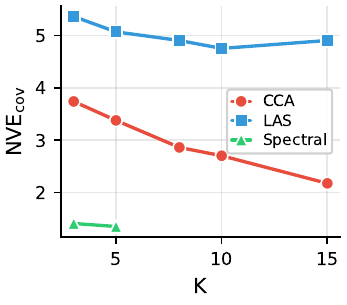}
        \caption{\texttt{\textbf{03_cdc_28}}}
    \end{subfigure}\hfill
    \begin{subfigure}{0.24\textwidth}
        \includegraphics[width=\linewidth]{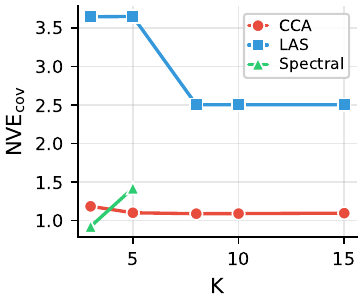}
        \caption{\texttt{\textbf{04_elutriation}}}
        \label{fig:ksweep_elutriation}
    \end{subfigure}
    
    \vspace{0.5em}
    
    \begin{subfigure}{0.24\textwidth}
        \includegraphics[width=\linewidth]{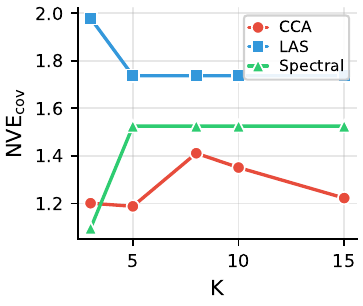}
        \caption{\texttt{\textbf{08_25mm_DTT}}}
    \end{subfigure}\hfill
    \begin{subfigure}{0.24\textwidth}
        \includegraphics[width=\linewidth]{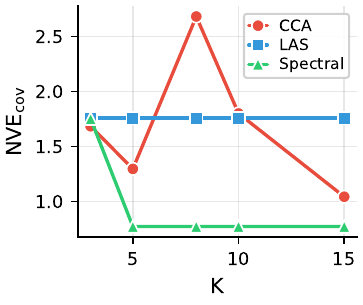}
        \caption{\texttt{\textbf{12_heat_shock_1}}}
    \end{subfigure}\hfill
    \begin{subfigure}{0.24\textwidth}
        \includegraphics[width=\linewidth]{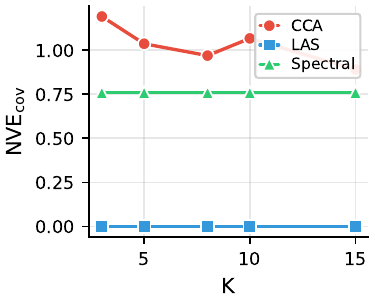}
        \caption{\texttt{\textbf{13_heat_shock_2}}}
        \label{fig:ksweep_heatshock2}
    \end{subfigure}\hfill
    \begin{subfigure}{0.24\textwidth}
        \includegraphics[width=\linewidth]{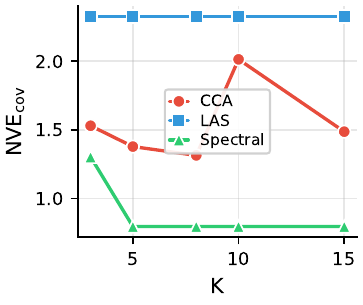}
        \caption{\texttt{\textbf{14_N\textsubscript{2}_depletion}}}
    \end{subfigure}
    
    \caption{Yeast benchmark $K$-sweeps scored by NVE\textsubscript{cov}.}
    \label{fig:yeast-ksweep}
\end{figure*}


The real-data argument also becomes sharper when the dataset-specific NVE\textsubscript{cov} sweeps are cited directly. In \texttt{02\_cdc\_15} (Fig.~\ref{fig:yeast-ksweep}(\subref{fig:ksweep_cdc15})), the NVE\textsubscript{cov} minimum occurs for Spectral at $K=3$, whereas ASR prefers CCA at $K=15$ and AvgVE prefers LAS at $K=15$. Thus, the metric choice changes both the preferred algorithm and the selected model complexity. In \texttt{04\_elutriation} (Fig.~\ref{fig:yeast-ksweep}(\subref{fig:ksweep_elutriation})), LAS performs best under AvgVE and NVE at $K=8$, but NVE\textsubscript{cov} instead selects Spectral at $K=3$, consistent with the fact that the LAS solution covers only about $5.2\%$ of the matrix while the selected Spectral solution covers about $43.3\%$. The main cautionary counterexample should also be pointed out explicitly. In \texttt{13\_heat\_shock\_2} (Fig.~\ref{fig:yeast-ksweep}(\subref{fig:ksweep_heatshock2})), LAS returns an effectively degenerate solution with coverage around $0.21\%$, yet all four metrics collapse to zero. This is not evidence against the usefulness of NVE or NVE\textsubscript{cov}, but it does show that a coverage penalty alone cannot rescue pathologically small outputs once the underlying VE term has already collapsed.



\subsection{Failure modes and caveats}

\begin{figure*}[t]
    \centering
    
    \begin{subfigure}{0.24\textwidth}
        \includegraphics[width=\linewidth]{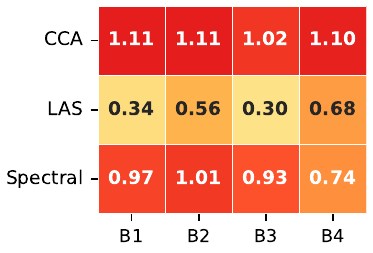}
        \caption{S1 additive (K=K\textsuperscript{*}=4)}
        \label{fig:heat-s1}
    \end{subfigure}\hfill
    \begin{subfigure}{0.24\textwidth}
        \includegraphics[width=\linewidth]{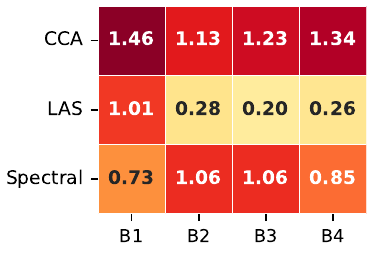}
        \caption{S2 scaling (K=K\textsuperscript{*}=4)}
    \end{subfigure}\hfill
    \begin{subfigure}{0.24\textwidth}
        \includegraphics[width=\linewidth]{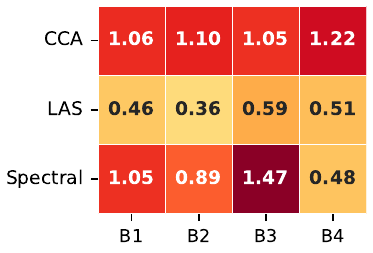}
        \caption{S3 mixed shift scale (K=K\textsuperscript{*}=4)}
    \end{subfigure}\hfill
    \begin{subfigure}{0.24\textwidth}
        \includegraphics[width=\linewidth]{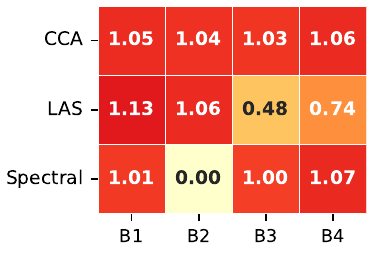}
        \caption{S4 size vs quality (K=K\textsuperscript{*}=4)}
    \end{subfigure}
    
    \vspace{0.5em}
    
    \begin{subfigure}{0.24\textwidth}
        \includegraphics[width=\linewidth]{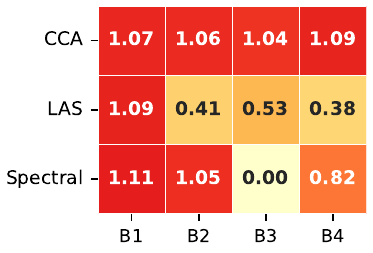}
        \caption{S5 overlap (K=K\textsuperscript{*}=4)}
    \end{subfigure}\hfill
    \begin{subfigure}{0.24\textwidth}
        \includegraphics[width=\linewidth]{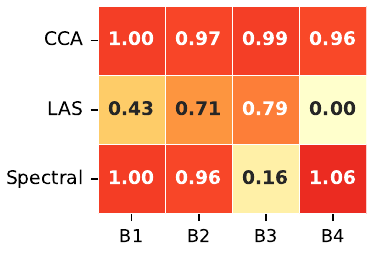}
        \caption{S6 thin columns (K=K\textsuperscript{*}=4)}
        \label{fig:heat-s6}
    \end{subfigure}\hfill
    \begin{subfigure}{0.24\textwidth}
        \includegraphics[width=\linewidth]{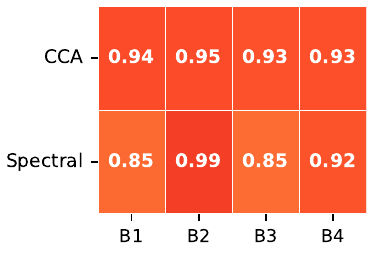}
        \caption{S7 identical clones (K=K\textsuperscript{*}=4)}
    \end{subfigure}\hfill
    \begin{subfigure}{0.24\textwidth}
        \includegraphics[width=\linewidth]{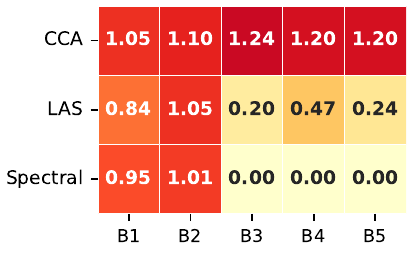}
        \caption{S8 noise ladder (K=K\textsuperscript{*}=5)}
    \end{subfigure}
    
    \vspace{0.5em}
    

    \centering
    \begin{subfigure}{0.24\textwidth}
        \includegraphics[width=\linewidth]{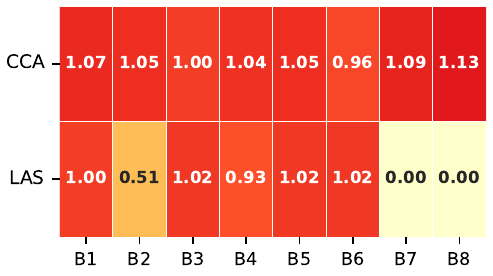}
        \caption{S9a coverage tiling (K=K\textsuperscript{*}=8)}
    \end{subfigure}\hspace{10pt}
    \begin{subfigure}{0.24\textwidth}
        \includegraphics[width=\linewidth]{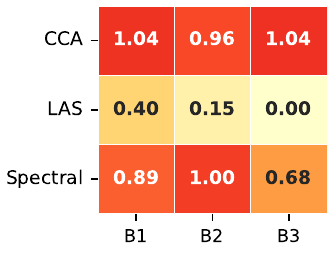}
        \caption{S9b cov. cherrypick (K=K\textsuperscript{*}=3)}
        \label{fig:heat-s9b}
    \end{subfigure}\hspace{20pt}
    \begin{subfigure}{0.1\textwidth}
        \includegraphics[width=\linewidth]{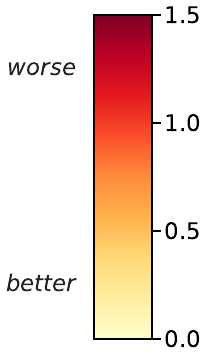}
        \caption{Reference}
        \label{fig:heat-ref}
    \end{subfigure}

    
    \caption{Per-bicluster NVE diagnostics on synthetic benchmark scenarios.}
    \label{fig:heatmap}
\end{figure*}

Although the main experiments suggest that NVE and NVE\textsubscript{cov} provide useful additional information, the results also show that these metrics should not be treated as automatically reliable in every situation. Some edge cases produce solutions that look very favourable numerically even though they are not especially useful from a biclustering point of view. For this reason, it is important to discuss the situations in which the proposed metrics can become misleading. Fig.~\ref{fig:heatmap} shows the local NVE contribution of each recovered bicluster, with Figs.~\ref{fig:heatmap}(\subref{fig:heat-s1})--~\ref{fig:heatmap}(\subref{fig:heat-s9b}) reporting the individual scenarios and Fig.~\ref{fig:heatmap}(\subref{fig:heat-ref}) showing the reference colourmap used for these heatmaps. In each heatmap, rows denote algorithms and columns denote biclusters $B_1,\ldots,B_{K^*}$ obtained at the scenario's ground-truth value of $K^*$. Each cell reports
$\operatorname{NVE}(B_k)=\operatorname{VE}(B_k)/\min_{\ell\neq k}\operatorname{VE}(B_{k\ell})$,
where $B_{k\ell}$ is the union super-bicluster formed from $B_k$ and $B_\ell$. Thus, lighter and smaller cells indicate biclusters that are internally coherent and remain well separated from their nearest merged alternative, whereas darker and larger cells identify noisy, redundant, or weakly separated local components. The figure should therefore be read as a diagnostic map of where an algorithm succeeds or fails within a scenario, rather than only as an aggregate ranking.

One clear example comes from the synthetic thin-column case. The per-bicluster diagnostic in Fig.~\ref{fig:heatmap}(\subref{fig:heat-s6}) shows that once a bicluster collapses to only two columns, the VE-based terms can become artificially small, which then propagates into AvgVE and the NVE family. The issue here is not genuine bicluster quality, but score degeneracy induced by an undersized submatrix. This is why the thin-column scenario should be cited directly when discussing the need for a minimum-size filter. A closely related problem appears in the yeast benchmark. Fig.~\ref{fig:yeast-ksweep}(\subref{fig:ksweep_heatshock2}) shows that \texttt{13\_heat\_shock\_2} contains a LAS solution whose score remains spuriously favourable despite negligible coverage. \textit{Taken together, Figs.~\ref{fig:heatmap}(\subref{fig:heat-s6}) and ~\ref{fig:yeast-ksweep}(\subref{fig:ksweep_heatshock2}) support a more precise caveat}: NVE-based measures are most informative when biclusters are large enough for VE to be numerically stable, and they should be reported together with basic size and coverage statistics whenever very small outputs are possible.


These edge cases suggest that an additional validity check may be needed before NVE-based scores are interpreted. One simple option would be to impose a minimum bicluster size, especially a minimum number of columns, before computing VE, NVE, or NVE\textsubscript{cov}. Another option would be to report NVE-family scores together with coverage and average bicluster dimensions, so that a very low score from a tiny solution is not mistaken for strong overall performance. We do not treat this as a failure of the main idea behind NVE, but rather as a reminder that evaluation metrics still depend on the kinds of solutions produced by the underlying algorithm.

Overall, these caveats do not remove the main findings of the paper. Instead, they help define the conditions under which the proposed metrics are most informative. NVE is most useful when the biclusters are large enough for VE to be stable and when the main question is whether the discovered biclusters are genuinely distinct. NVE\textsubscript{cov} is most useful when low-coverage cherry-picking is a realistic concern. However, in the presence of very small or degenerate biclusters, additional filtering or reporting rules are likely to be necessary.


\subsection{Relation to NSR}
The closest prior internal measure to NVE is the normalized squared residue (NSR) of Lee et al.~\cite{koreanPaper}. NSR and NVE use the same general normalization idea: the coherence of each bicluster is compared with the coherence of a super-bicluster formed by merging it with another bicluster. Their difference is the elementary coherence functional. NSR is defined from the mean squared residue (Eq.~\eqref{eq:nsr}), whereas NVE replaces $\operatorname{MSR}(\cdot)$ by $\operatorname{VE}(\cdot)$. Thus, NSR is best viewed as the MSR-based member of the same super-bicluster-normalization family, rather than as an independent pattern-consistency measure.




This distinction is important mathematically. For an additive bicluster $a_{ij}=\mu+\alpha_i+\beta_j+\epsilon_{ij}$,
MSR measures the residual noise variance and is well matched to the assumed model. However, for a noiseless multiplicative bicluster $a_{ij}=s_i p_j$,
the Cheng--Church residue is
\[
a_{ij}-a_{iJ}-a_{Ij}+a_{IJ}
=
(s_i-\bar{s})(p_j-\bar{p}),
\]
and hence
\[
\operatorname{MSR}(B)
=
\left(\frac{1}{|I|}\sum_{i\in I}(s_i-\bar{s})^2\right)
\left(\frac{1}{|J|}\sum_{j\in J}(p_j-\bar{p})^2\right),
\]
which is generally nonzero even when the bicluster follows a perfect scaling pattern. NSR cannot remove this model bias, because both its numerator and denominator are still computed from MSR. NVE was introduced precisely to retain the super-bicluster separability idea while using the VE coherence model, which is more appropriate for standardized behavioural patterns.

To verify that this distinction is not merely formal, we computed NSR only on a small diagnostic subset of the planted synthetic biclusters, without rerunning the full algorithmic benchmark. The results in Table~\ref{tab:nsr_diagnostic} show the expected behavior. NSR and NVE both identify the identical-clone case as poorly separated, confirming that NSR is a meaningful predecessor for additive redundancy. However, NSR is much more strongly affected by the scaling scenario because it inherits MSR's additive-model bias. It also prefers the low-coverage cherry-picked solution over the broader tiling solution, whereas NVE$_{\mathrm{cov}}$ reverses this ranking by design.

\begin{table}
\centering
\caption{Targeted NSR diagnostic on planted synthetic biclusters.}
\label{tab:nsr_diagnostic}
\resizebox{\columnwidth}{!}{
\begin{tabular}{lcccccc}
\hline
\textbf{Scenario} & \textbf{ASR} & \textbf{AvgVE} & \textbf{NSR} & \textbf{NVE} & \textbf{NVE\textsubscript{cov}} & \textbf{Coverage} \\
\hline
S1 additive & 0.0861 & 0.1548 & 0.0372 & 0.2587 & 0.7292 & 0.1258 \\
S2 scaling & 1.4897 & 0.1561 & 0.2443 & 0.2231 & 0.6291 & 0.1258 \\
S7 identical clones & 0.0958 & 0.2339 & 1.0366 & 1.0973 & 3.2126 & 0.1167 \\
S9a coverage tiling & 0.2041 & 0.3497 & 0.1349 & 0.4709 & 1.6312 & 0.0833 \\
S9b coverage cherry-pick & 0.0517 & 0.2331 & 0.0192 & 0.2716 & 2.7163 & 0.0100 \\
\hline
\end{tabular}
}
\end{table}

For this reason, we do not treat NSR as a full baseline throughout all experiments. The main empirical question is whether VE-based normalization and coverage adjustment add information beyond standard coherence summaries. NSR is therefore used here as a conceptual and diagnostic reference: it validates the super-bicluster normalization principle in the additive setting, while also illustrating why the proposed VE-based and coverage-adjusted variants are needed.
\section{Conclusion}

This paper examined whether Normalised Virtual Error (NVE) and its coverage-aware variant NVE\textsubscript{cov} provide evaluation information beyond that already captured by existing internal biclustering measures. The central motivation was that commonly used measures such as ASR, MSR, and AvgVE mainly assess within-bicluster coherence, but do not explicitly evaluate whether the extracted biclusters are genuinely distinct from one another or whether they explain a meaningful portion of the data matrix. To address this gap, we studied NVE as a VE-based criterion that incorporates relative distinctness through comparison with merged super-biclusters, and NVE\textsubscript{cov} as a further extension that penalises low coverage at the solution level.

Taken together, the experiments support three main conclusions. First, NVE captures information about separation and redundancy that is not fully reflected by coherence-based measures. This was most clearly seen in the controlled synthetic datasets, where overlapping or near-duplicate biclusters could still look favourable under coherence-based scoring but were penalised more clearly by NVE. Second, NVE\textsubscript{cov} adds a further evaluation component by discouraging solutions that achieve low coherence error while covering only a very small fraction of the matrix. In both the synthetic benchmarks and the yeast benchmark \(K\)-sweeps, this adjustment could change the preferred algorithm and, in the real-data setting, the selected model complexity. Third, these effects show that NVE-based criteria do not merely rescale existing coherence measures: they make separability and representativeness explicit, and can therefore lead to different judgements about biclustering quality.

At the same time, the results also make clear that NVE-based evaluation should not be interpreted uncritically. When biclusters become extremely small or degenerate, especially in thin-column settings, VE-based quantities can become artificially favourable. The yeast experiments showed a related failure mode in which a negligible-coverage solution could still receive extremely strong scores. For this reason, we do not claim that NVE or NVE\textsubscript{cov} eliminates the need for basic sanity checks on bicluster size and coverage. Rather, the proposed measures are most useful when applied to solutions for which the underlying VE computation is numerically meaningful, and when they are reported together with simple structural diagnostics such as average bicluster dimensions and coverage.

Overall, the main conclusion is that NVE and especially NVE\textsubscript{cov} contribute useful additional internal validation information for biclustering and co-clustering. NVE adds a notion of relative separability and redundancy that is absent from standard coherence-only measures, while NVE\textsubscript{cov} adds a notion of representativeness by discouraging narrow, low-coverage solutions. Together, these properties make the proposed measures useful not only for scoring biclustering solutions, but also for practical model selection. Future work should study more robust VE-based formulations for very small biclusters, investigate alternative coverage and overlap regularisations, and evaluate the proposed criteria across a broader range of biclustering algorithms and application domains.

\bibliography{References}
\bibliographystyle{IEEEtran}

\end{document}